\documentclass[10pt,twocolumn,letterpaper]{article}

\usepackage{cvpr}
\usepackage{times}
\usepackage{epsfig}
\usepackage{graphicx}
\usepackage{amsmath}
\usepackage{amssymb}
\usepackage{pifont}
\usepackage{algorithm}
\usepackage{algorithmic}
\newcommand{\cmark}{\ding{51}}
\newcommand{\xmark}{\ding{55}}

\usepackage[breaklinks=true,bookmarks=false]{hyperref}

\cvprfinalcopy 

\def\cvprPaperID{****} 

\begin{document}

\title{Improving Annotation for 3D Pose Dataset of Fine-Grained Object Categories}

\author{Yaming Wang\\
{\tt\small wym@umiacs.umd.edu}
\and
Xiao Tan\\
{\tt\small tanxiao01@baidu.com}
\and
Yi Yang\\
{\tt \small yangyi05@baidu.com}
\and
Ziyu Li\\
{\tt \small ziyuli@baidu.com}
\and
Xiao Liu\\
{\tt \small liuxiao12@baidu.com}
\and
Feng Zhou\\
{\tt \small zhoufeng09@baidu.com}
\and
Larry S. Davis\\
{\tt \small lsd@umiacs.umd.edu}
}

\maketitle

\begin{abstract}
Existing 3D pose datasets of object categories are limited to generic object types and lack of fine-grained information.
In this work, we introduce a new large-scale dataset that consists of 409 fine-grained categories and 31,881 images with accurate 3D pose annotation.
Specifically, we augment three existing fine-grained object recognition datasets (StanfordCars, CompCars and FGVC-Aircraft) by finding a specific 3D model for each sub-category from ShapeNet and manually annotating each 2D image by adjusting a full set of 7 continuous perspective parameters.
Since the fine-grained shapes allow 3D models to better fit the images, we further improve the annotation quality by initializing from the human annotation and conducting local search of the pose parameters with the objective of maximizing the IoUs between the projected mask and the segmentation reference estimated from state-of-the-art deep Convolutional Neural Networks (CNNs).
We provide a full statistics of the annotations with qualitative and quantitative comparisons suggesting that our dataset can be a complementary source for studying 3D pose estimation.
The dataset can be downloaded at 
\url{http://users.umiacs.umd.edu/~wym/3dpose.html}.
\end{abstract}

\section{Introduction}

\setlength{\tabcolsep}{1pt}
\begin{figure}[t]
  \footnotesize
  \centering
  \begin{tabular}{c c c}
    \includegraphics[width=0.32\linewidth,height=0.22\linewidth]{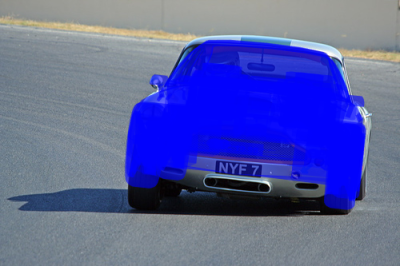} &
    \includegraphics[width=0.32\linewidth,height=0.22\linewidth]{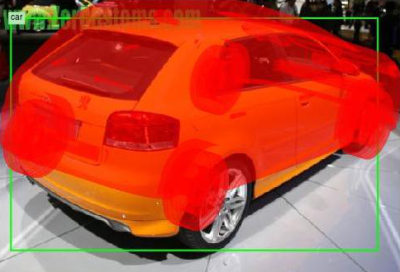} & 
    \includegraphics[width=0.32\linewidth,height=0.22\linewidth]{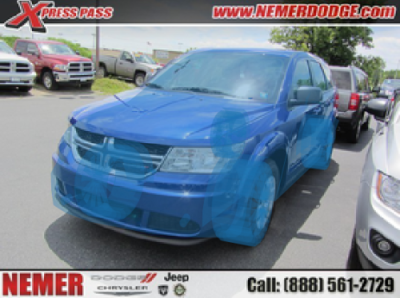} \\
    Pascal3D+ & ObjectNet3D & StanfordCars3D (Ours) \\
    \includegraphics[width=0.32\linewidth,height=0.22\linewidth]{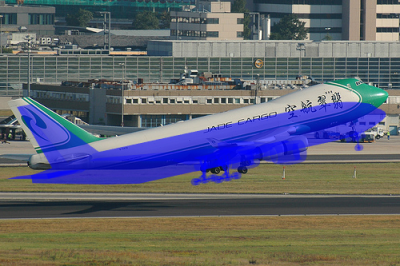} &
    \includegraphics[width=0.32\linewidth,height=0.22\linewidth]{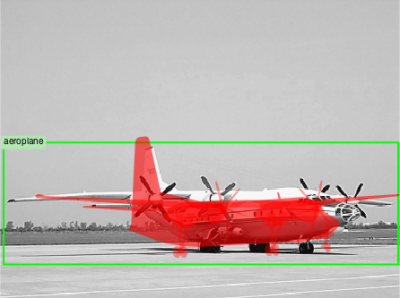} & 
    \includegraphics[width=0.32\linewidth,height=0.22\linewidth]{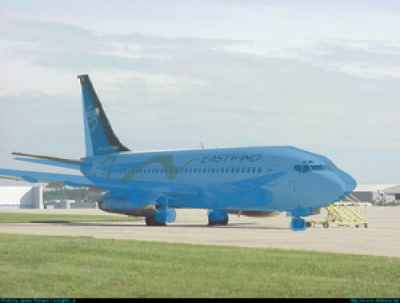} \\
    Pascal3D+ & ObjectNet3D & FGVC-Aircraft3D (Ours) \\   
  \end{tabular}
  \caption{While both Pascal3D+ and ObjectNet3D contain more complicated scenarios with more generic categories for 3D pose estimation, we provide more accurate pose annotations on a large set of fine-grained object classes as a complementary source for studying 3D pose estimation.}
  \label{fig:dataset_compare_simple}
\end{figure}

In the past few years, the fast-pacing progress of generic image recognition on ImageNet~\cite{krizhevsky2012imagenet} has drawn increasing attention in classifying fine-grained object categories~\cite{krause2016unreasonable,van2017inaturalist}, \eg bird species~\cite{wah2011cub}, car makes and models~\cite{krause20133d}.
However, simply recognizing object labels is still far from solving many industrial problems where we need a deeper understanding of other attributes of the objects~\cite{lim2013parsing}.
On the other hand, estimating 3D object pose from a single 2D image is an indispensable step in various practical applications, such as vehicle damage detection~\cite{jayawardena2013image},
novel view synthesis~\cite{zhou2016view,park2017transformation}, grasp planning~\cite{varley2017shape} and autonomous driving~\cite{chen2016monocular}.
In this work, we introduce the problem of estimating 3D pose for fine-grained objects from monocular images.
We believe this will become an important component in broader tasks, contributing to both fine-grained object recognition and 3D object pose estimation.

To address this task, collecting suitable data is of vital importance. 
However, due to the expensive annotation cost, most existing 3D pose datasets only provide accurate ground truth annotations for a few object classes and the number of instances associated to each category is quite small~\cite{ozuysal2009pose}. 
Although there are two large scale pose datasets, Pascal3D+~\cite{xiang2014beyond} and ObjectNet3D~\cite{xiang2016objectnet3d}, both of them are collected for generic object types and there is still no large-scale 3D pose dataset for fine-grained object categories.
Moreover, these datasets are lack of accurate pose information, since different objects in one hyper class (\ie, cars) are only matched with a few generic 3D shapes, leading to a high projection error that affects human annotators to find the accurate pose, as demonstrated in Figure~\ref{fig:dataset_compare_simple}.

In this work, we introduce a new benchmark pose estimation dataset for fine-grained object categories.
Specifically, we augment three existing fine-grained recognition datasets, StanfordCars~\cite{krause20133d}, CompCars~\cite{yang2015large} and FGVC-Aircraft~\cite{maji2013fine}, with two types of useful 3D information: (1) for each object in the image, we manually annotate the full perspective projection represented by 7 continuous pose parameters; (2) we provide an accurate match of the computer aided design (CAD) model for each fine-grained object category. 
The resulting augmented dataset consists of more than 30,000 images for over 400 fine-grained object categories.
Table~\ref{table:dataset} shows the general statistics of our dataset.

\begin{figure}[t]
  \scriptsize
  \centering
  \begin{tabular}{c c c}
    \includegraphics[width=0.32\linewidth,height=0.20\linewidth]{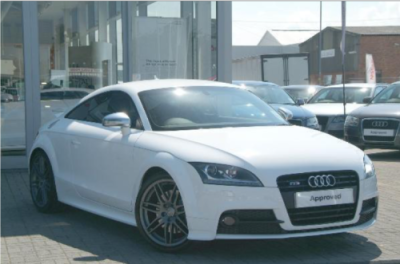} &
    \includegraphics[width=0.32\linewidth,height=0.20\linewidth]{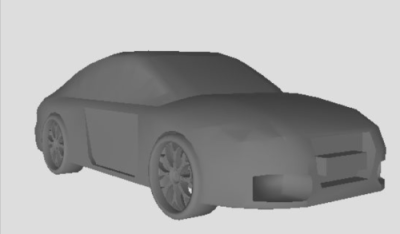} &
    \includegraphics[width=0.32\linewidth,height=0.20\linewidth]{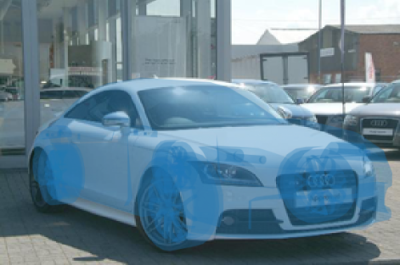} \\
    Fine-grained 2D Image & Fine-grained 3D Model & Initial Pose by Human \\
    \includegraphics[width=0.32\linewidth,height=0.20\linewidth]{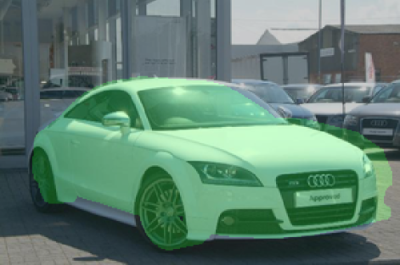} &
    \includegraphics[width=0.32\linewidth,height=0.20\linewidth]{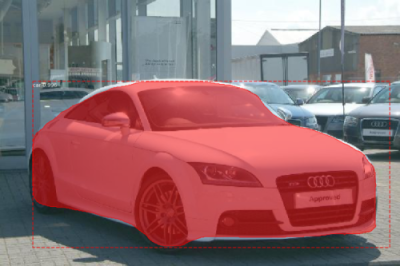} &
    \includegraphics[width=0.32\linewidth,height=0.20\linewidth]{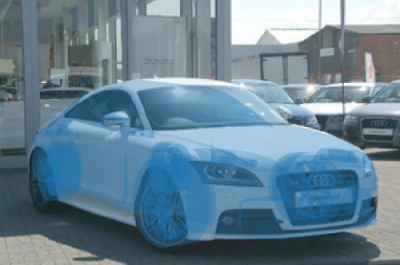} \\
    Initial 2D Segmentation & Segmentation Reference & Final Adjusted Pose \\
  \end{tabular}
  \caption{For an image with a fine-grained category (Top left), we first find its corresponding fine-grained 3D model (Top middle) and manually annotate its rough pose (Top right). Since the problem is to estimate the object pose such that the projection of the 3D model aligns with the image as well as possible, we further optimize the segmentation overlap between the projected 2D mask (Bottom left) and the ``groundtruth'' mask (Bottom middle) estimated from state-of-the-art CNN models to obtain the final 3D pose (Bottom right).}
  \label{fig:splash}
\end{figure}

To the best of our knowledge, our dataset is the very first one which employs fine-grained category aware 3D models in pose annotation.
To fully utilize the valuable fine-grained information, we further develop an automatic pose refinement mechanism to improve over the human annotations.
Thanks to the fine-grained shapes, an accurate pose parameter also leads to the optimal segmentation overlap between the projected 2D mask from the 3D model and the target object ground truth segmentation.
We hence conduct a local greedy search over the 7 full perspective pose parameters, initialized from the human annotation, to maximize the segmentation overlap objective.
To avoid effort on segmentation annotation, we utilize state-of-the-art image segmentation models including both Mask R-CNN~\cite{he2017mask} and DeepLab v3+~\cite{deeplabv3plus2018} to obtain the as-accurate-as-possible segmentation reference.
This process significantly improves our annotation quality.
Figure~\ref{fig:splash} illustrates this process. 

In summary, our contribution is three-fold. 
(1) We collect a new large-scale 3D pose dataset for fine-grained objects with more accurate annotations, which can be viewed as a complementary source to the existing pose dataset.
(2) Our pose annotation contains a full perspective model parameters including the camera focal length, which is a more challenging benchmark for developing algorithms beyond only estimating viewpoint angles (azimuth)~\cite{ghodrati20142d} or recovering the rotation matrices~\cite{mahendran20173d}. 
(3) We propose a simple but effective way to automatically refine the pose annotation based on the segmentation cues. With the corresponding 3D fine-grained model, this method can automatically refine object pose while significantly alleviating the human label effort.

\begin{table}
\begin{center}
\small
\begin{tabular}{|c|c|c|c|c|}
\hline
Dataset & \# class & \# image & annotation & fine-grained \\
\hline
3D Object~\cite{savarese20073d} & 10 & 6,675 & discretized view & \xmark \\
EPFL Car~\cite{ozuysal2009pose} & 1 & 2,299 & continuous view & \xmark \\
IKEA~\cite{lim2013parsing} & 11 & 759 & 2d-3d alignment & \xmark \\
Pascal3D+~\cite{xiang2014beyond} & 12 & 30,899 & 2d-3d alignment & \xmark \\
ObjectNet3D~\cite{xiang2016objectnet3d} & 100 & 90,127 & 2d-3d alignment & \xmark \\
\hline
StanfordCars3D & 196 & 16,185 & 2d-3d alignment & \cmark \\
CompCars3D & 113 & 5,696 & 2d-3d alignment & \cmark \\
FGVC-Aircraft3D & 100 & 10,000 & 2d-3d alignment & \cmark \\
Total (Ours) & 409 & 31,881 & 2d-3d alignment & \cmark \\
\hline
\end{tabular}
\caption{Comparison between our 3D pose estimation dataset (StanfordCars3D + CompCars3D + FGVC-Aircraft3D) and other benchmark datasets. Our dataset can be viewed as a complementary source to the existing large scale 3D pose dataset (Pascal3D+ and ObjectNet3D) with a different focus on more intra-class categories and fine-grained details.}
\label{table:dataset}
\end{center}
\end{table}

\setlength{\tabcolsep}{1pt}

\begin{figure*}[t]
\centering
\begin{tabular}{c}
\includegraphics[width=0.9\textwidth]{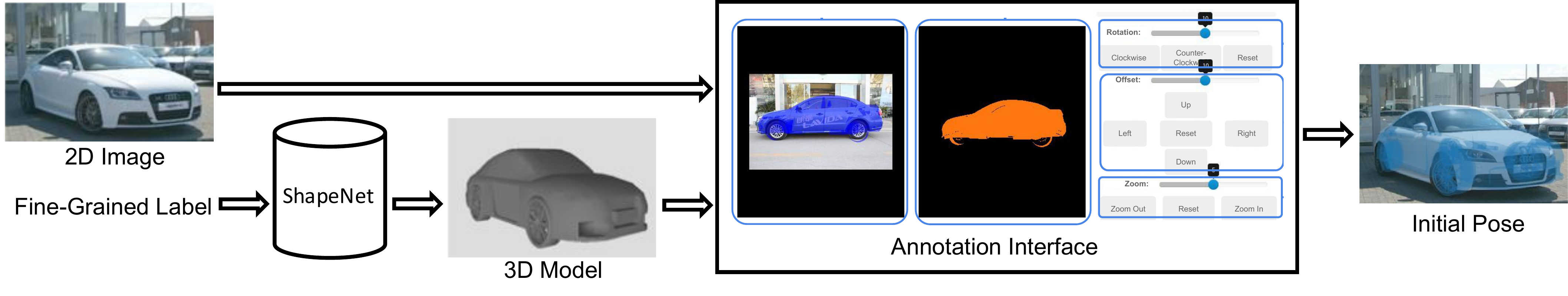} \\
(1) human pose annotation \\
\includegraphics[width=0.9\textwidth]{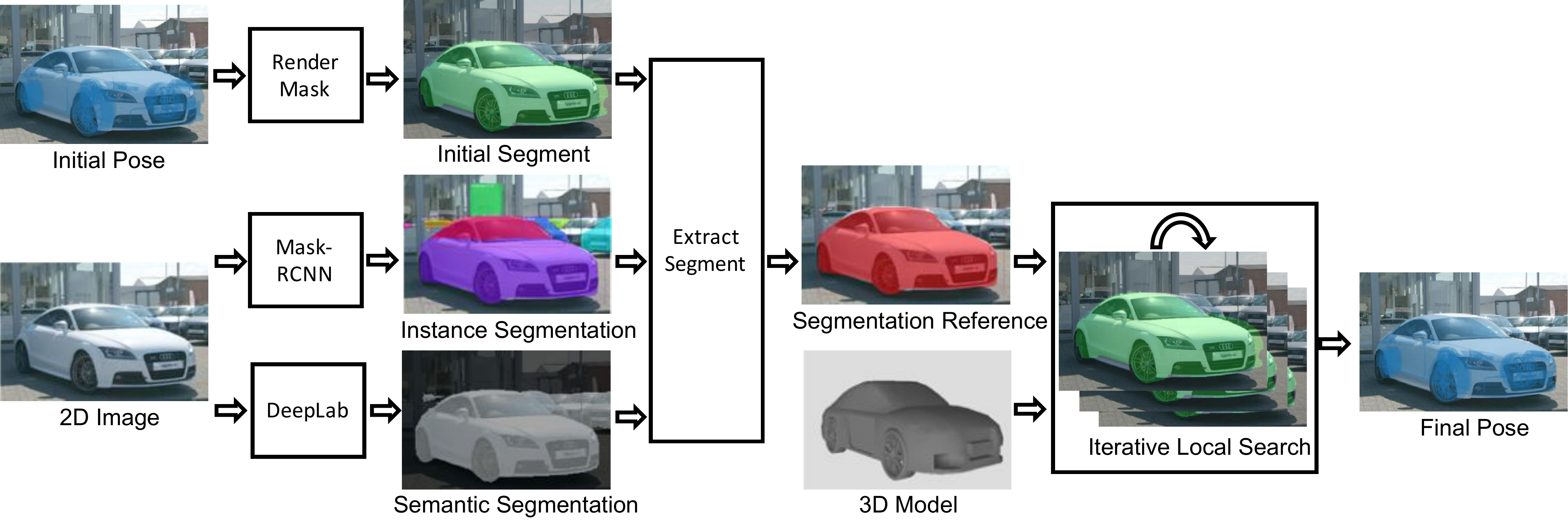} \\
(2) segmentation based pose refinement \\
\end{tabular}
\caption{An overview of our whole annotation framework which includes two parts: (1) human initial pose annotation, and (2) segmentation based pose refinement. The human annotation provides a strong initialization for the second-stage pose refinement, hence we only need to conduct local search to adjust the pose.}
\label{fig:framework}
\end{figure*}

\section{Related Work}
\textbf{3D Pose Estimation Dataset.} 
Due to the 3D ambiguity from 2D images and heavy annotation cost, earlier object pose datasets are limited not only in their dataset scales but also in the types of annotation they covered.
Table~\ref{table:dataset} provides a quantitative comparison between our dataset and previous ones.
For example, 3D Object dataset~\cite{savarese20073d} only provides viewpoint annotation for 10 object classes with 10 instances for each class.
EPFL Car dataset~\cite{ozuysal2009pose} consists of 2,299 images of 20 car instances captured at multiple azimuth angles.
Moreover, the other parameters including elevation and distance are kept almost the same for all the instances in order to simplify the problem~\cite{ozuysal2009pose}.
Pascal3D+~\cite{xiang2014beyond} is perhaps the first large-scale 3D pose dataset for generic object categories, with 30,899 images from 12 different classes of the Pascal VOC dataset~\cite{everingham2010pascal}.
Recently, ObjectNet3D~\cite{xiang2016objectnet3d} further extends the dataset scale to 90,127 images of 100 categories.
Both Pascal3D+ and ObjectNe3D assume a camera model with 6 parameters to annotate.
However, different images in one hyper class (\ie, cars) are usually matched with a few coarse 3D CAD models, thereby the projection error might be large due to the lack of accurate CAD models in some cases.
Being aware of these problems, we therefore project fine-grained CAD models to match objects in the 2D images.
In addition, our datasets surpass most of previous ones in both scales of images and classes.

\textbf{Fine-Grained Recognition Dataset.} Fine-grained recognition refers to the task of distinguishing sub-ordinate categories~\cite{wah2011cub,krause20133d,van2017inaturalist}. 
In earlier works, 3D information is a common source to gain recognition performance improvement~\cite{zia2013detailed,xiang2015data,mottaghi2015coarse,sochor2016boxcars}.
As deep learning prevails and fine-grained datasets become larger, the effect of 3D information on recognition diminishes~\cite{lin2015bilinear,krause2016unreasonable}. 
Recently, \cite{sochor2016boxcars} incorporate 3D bounding box into deep framework when images of cars are taken from a fixed camera. 
On the other hand, almost all existing fine-grained datasets are lack of 3D pose labels or 3D shape information~\cite{krause20133d}, and pose estimation for fine-grained object categories are not well-studied. 
Our work fills this gap by annotating poses and matching CAD models on three existing popular fine-grained recognition datasets.

\textbf{3D Model Dataset.}
Similar to~\cite{xiang2016objectnet3d}, we adopt the 2d-3d alignment method to annotate object poses, 
Annotating in such a way requires a source for accessing accurate 3D models of objects.
Luckily, there has been substantial growth in the number of of 3D models available online over the last decade~\cite{chen2009benchmark,chen2012schelling,kim2013learning,li2014shrec} with well-known repositories like the Princeton Shape Benchmark~\cite{shilane2004princeton} which contains around 1,800 3D models grouped into 90 categories.
In this work, we use ShapeNet~\cite{chang2015shapenet}, the so far largest 3D CAD model database which has indexed more than 3,000,000 models, with 220,000 models out of which are classified into 3,135 categories including various object types such as cars, airplanes, bicyles, etc.
The large amount of 3D models allow us to find an exact model to many of the objects in the natural images.
For example, the car category, ShapeNet provides 183,533 models for the car category and 114,045 models for the airplane category.
Note that although we only annotate three fine-grained datasets, our annotation framework can be continued to apply to building more 3D pose dataset, thanks to larger-scale datasets like ShapeNet~\cite{chang2015shapenet} and iNaturalist~\cite{van2017inaturalist}.

\section{Dataset Construction}

Building our 3D pose dataset involves two main processes: (1) human pose annotation, and (2) segmentation based pose refinement.
Figure \ref{fig:framework} illustrates the whole process.
Our human pose annotation process is similar to ObjectNet3D~\cite{xiang2016objectnet3d} but requires more effort on selecting finer 3D models.
We first select the most appropriate 3D car model from ShapeNet~\cite{chang2015shapenet} for each object in the fine-grained image dataset.
We then obtain the pose parameters by asking the annotators to align the projection of the 3D model to the corresponding image using our designed interface.

Although human can initiate the pose annotation with reasonably high efficiency and accuracy, we find it hard for them to adjust the fine detailed poses.
Our second-stage segmentation based pose refinement further adjusts the pose parameters by performing a local greedy search initialized from the human annotation.
We discuss the details of each process in the next subsections.

\begin{figure*}[t]
\centering
\includegraphics[width=0.9\textwidth]{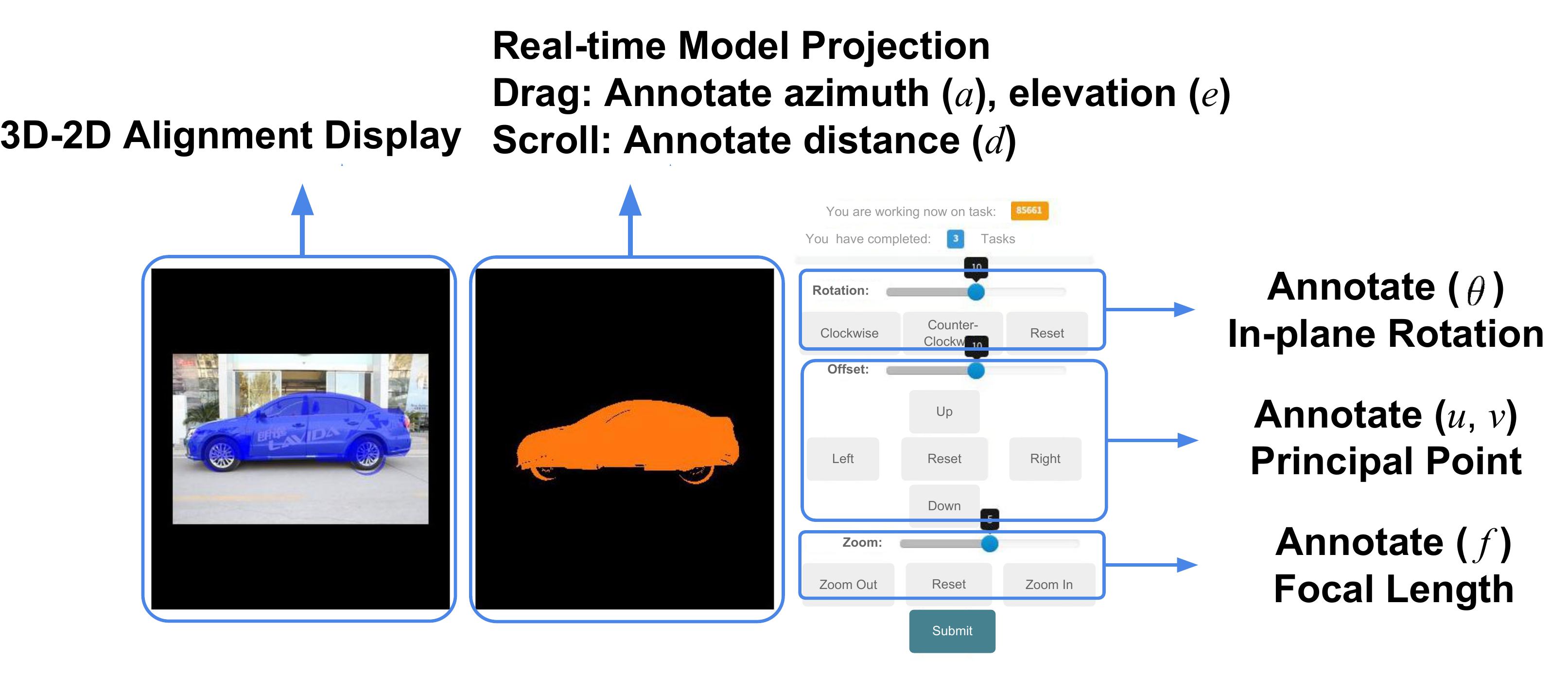}
\vspace{-10pt}
\caption{An overview of our annotation interface. 
Our annotation tool renders the projected 2D mask onto the image in real time to facilitate the annotators to better adjust pose parameters.}
\label{fig:annotation_interface}
\end{figure*}

\subsection{3D Models}
We build three fine-grained 3D pose datasets. Each dataset consists of two parts: 2D images and 3D models.
The 2D images are collected from StanfordCars \cite{krause20133d}, CompCars \cite{yang2015large} and FGVC-Aircraft \cite{maji2013fine} respectively.
Unlike Pascal3D+ \cite{xiang2014beyond} and ObjectNet3D \cite{xiang2016objectnet3d}, the target objects in most images are non-occluded and easy to identify.
In order to distinguish between fine-grained categories, we adopt a distinct model for each category.
Thanks to ShapeNet \cite{chang2015shapenet}, a large number of 3D models for fine-grained objects are available with make/model names in their meta data, which are used to find the corresponding 3D model given an image category name.
If there is no exact match between a category name and the meta data, we manually select a visually similar 3D model for that category. 
For StanfordCars, we annotate images for all 196 categories, where 148 categories have exact matched 3D models. 
For CompCars, we only include 113 categories with matched 3D models in ShapeNet.
For FGVC-Aircraft, we annotate images for all 100 categories with more than 70 matched models.
To the best of our knowledge, our dataset is the very first one which employs fine-grained category aware 3D models in 3D pose estimation.

\subsection{Camera Model}
The world coordinate system is defined in accordance with the 3D model coordinate system. 
In this case, a point $\bf X$ on a 3D model is projected onto a point $\bf x$ on a 2D image:
\begin{equation}\label{eq1}
{\bf x} = {\cal{P}}{\bf X},
\end{equation}
via a perspective projection matrix:
\begin{equation}\label{eq2}
{\cal P} = K\left[ {R|T} \right],
\end{equation}
where $K$ denotes the intrinsic parameter matrix:
\begin{equation}\label{eq3}
K = \left[ {\begin{array}{*{20}{c}}
f&0&{{u}}\\
0&f&{{v}}\\
0&0&1
\end{array}} \right],
\end{equation}
and $R$ encodes a $3 \times 3$ rotation matrix between the world and camera coordinate systems, parameterized by three angles, \ie, elevation $e$, azimuth $a$ and in-plane rotation $\theta$.
We assume that the camera is always facing towards the origin of the 3D model. Hence the translation $T = [0, 0, d]^{\rm T}$ is only defined up to the model depth $d$, the distance between the origins of the two coordinate systems, and the principal point $(u, v)$ is the projection of the origin of world coordinate system on the image. 
As a result, our model has 7 continuous parameters in total: camera focal length $f$, principal point location $u$, $v$, azimuth $a$, elevation $e$, in-plane rotation $\theta$ and model depth $d$. 
Note that, since the images are collected online, the annotated intrinsic parameters ($u$, $v$ and $f$) are approximations. Compared to previous datasets \cite{xiang2014beyond,xiang2016objectnet3d} with 6 parameters ($f$ fixed), our camera model considers both the camera focal length $f$ and object depth $d$ in a full perspective projection for finer 2D-3D alignment, which allows for a more flexible pose adjustment and a better shape matching.

\subsection{2D-3D Alignment}

We annotate 3D pose information for all 2D images through crowd-sourcing. 
To facilitate the annotation process, we develop an annotation tool illustrated in Figure \ref{fig:annotation_interface}. 
For each image during annotation, we choose the 3D model according to the fine-grained label given beforehand. 
We then ask the annotators to adjust the 7 parameters so that the projected 3D model is aligned with the target object in the 2D image. 
This process can be roughly summarized as follows: (1) shift the 3D model such that the center of the model (the origin of the world coordinate system) is roughly aligned with the center of the target object in the 2D image; (2) rotate the model to the same orientation as the target object in the 2D image; (3) adjust the model depth $d$ and camera focal length $f$ to match the size of the target object in the 2D image. 
Some finer adjustment might be applied after the three main steps. 
In this way we annotate all 7 parameters across the whole dataset. 
On average, each image takes approximately 1 minute to annotate by an experienced annotator. 
To ensure the quality, after one round of annotation across the whole dataset, we perform quality check and let the annotators do a second round revision for the unqualified examples.

\subsection{Segmentation Based Pose Refinement}

\begin{algorithm}[t]
\caption{Iterative local pose search algorithm:}
\begin{algorithmic}[1]
\renewcommand{\algorithmicrequire}{\textbf{Input:}}
\renewcommand{\algorithmicensure}{\textbf{Output:}}
\REQUIRE{3D model $\cal{M}$, Human pose annotation $\boldsymbol{p}_0$, segmentation reference $s^*$, 2D mask generator $S(\boldsymbol{p}, {\cal{M}})$, segmentation evaluation function $IoU(s_1, s_2)$, pose parameter update unit $\boldsymbol{\epsilon}$, update step size $\alpha$.
}
\ENSURE{Optimized pose parameter $\boldsymbol{p}^*$.}
\FOR{each image with segmentation reference $s^*$}
\STATE {Initialize pose parameters $\boldsymbol{p} = \boldsymbol{p}_0$.}
\STATE {Initialize 2D mask $s = S(\boldsymbol{p}, {\cal{M}})$}
\STATE {Initialize $iou = IoU(s, s^*)$}
\REPEAT
\STATE Update $iou_{last} = iou$.
\FOR{each dimension $i$ in $\boldsymbol{p}$}
\STATE Update $\boldsymbol{p}_i^+ = \boldsymbol{p}_i + \alpha \boldsymbol{\epsilon}_i$ 
\STATE Update $\boldsymbol{p}_i^- = \boldsymbol{p}_i - \alpha \boldsymbol{\epsilon}_i$
\STATE Render new 2D mask $s^+ = S(\boldsymbol{p}^+, {\cal{M}})$
\STATE Render new 2D mask $s^- = S(\boldsymbol{p}^-, {\cal{M}})$
\STATE Update $iou^+ = IoU(s^+, s^*)$
\STATE Update $iou^- = IoU(s^-, s^*)$
\STATE Update $iou = \max (iou, iou^+, iou^-)$
\STATE Update $\boldsymbol{p} = \arg\max (iou, iou^+, iou^-)$
\ENDFOR
\IF {$iou == iou_{last}$}
\STATE Update $\alpha = \alpha / 2$ 
\IF {$\alpha <= threshold $}
\STATE Set as convergence.
\ENDIF
\ELSE
\STATE Continue.
\ENDIF
\UNTIL{converge}
\ENDFOR
\end{algorithmic}
\label{alg:local_search}
\end{algorithm}

\setlength{\tabcolsep}{1pt}
\begin{figure}[t]
  \small
  \centering
  \begin{tabular}{c c}
    \includegraphics[width=0.40\linewidth,height=0.30\linewidth]{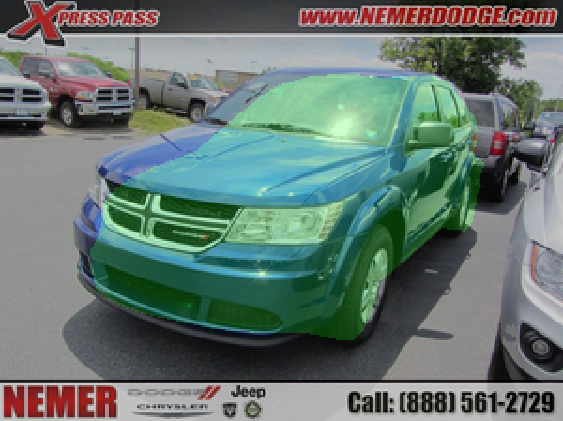} &
    \includegraphics[width=0.40\linewidth,height=0.30\linewidth]{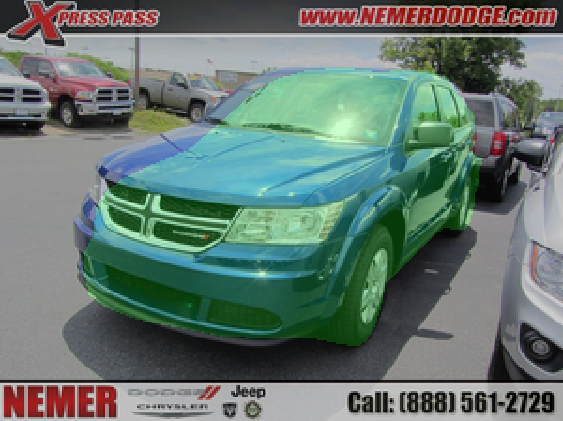} \\
    Initial Pose by Human & Iteration 1 \\
    \includegraphics[width=0.40\linewidth,height=0.30\linewidth]{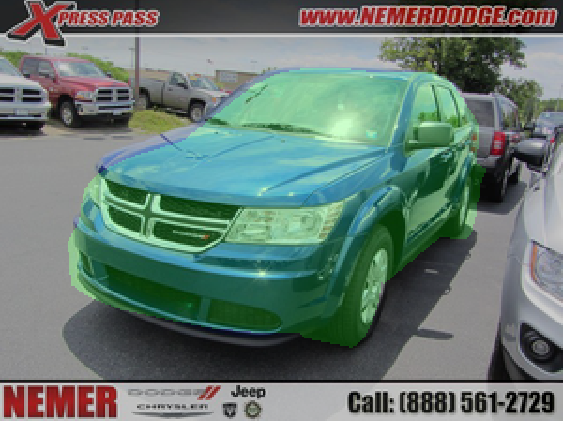} &
\includegraphics[width=0.40\linewidth,height=0.30\linewidth]{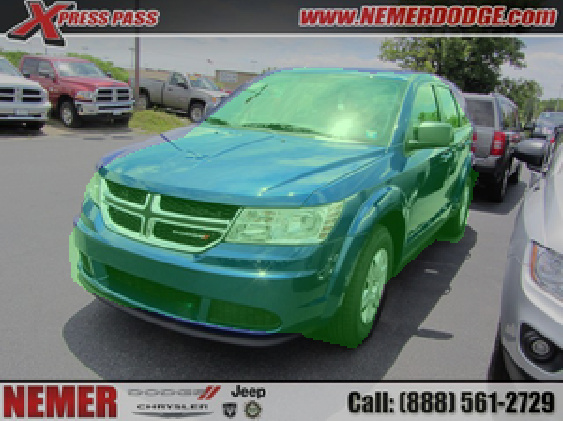} \\
    Iteration 2 & Final Pose \\
  \end{tabular}
  \caption{Iterative local greedy search for the fine detailed pose, initialized from human annotation. The green highlights are the 2D masks projected by the 3D model during pose optimization.}
  \label{fig:local_search}
\end{figure}

Although human annotators already provide reasonably accurate annotation in the first stage, we notice that there are still potential rooms to further improve the annotation quality.
This is because humans are good at providing a strong initial pose estimate but finetuning the detailed pose parameters is a very annoying thing to them.
Realizing that ultimately the problem is to estimate the object pose such that the projection of the 3D model aligns with the image as well as possible, we design a simple but effective iterative local greedy search algorithm to automatically adjust pose parameters by maximimzing
\begin{equation}
\max_{\boldsymbol{p}} J(\boldsymbol{p}) = IoU(S(\boldsymbol{p}, {\cal{M}}), s^*) 
\end{equation}
where $s^*$ is the 2D object segmentation reference and $S(\boldsymbol{p}, {\cal{M}})$ maps a 3D model $\cal{M}$ to a 2D mask according to the pose parameter $\boldsymbol{p} = (a, e, \theta, d, f, u, v)$.

The algorithm aims to finetune the 7 pose parameters to maximize the segmentation overlap between the projected 2D mask from the 3D model and the segmentation reference.
We use the traditional {\em intersection over union} as the segmentation overlapping criterion.
The algorithm greedily updates pose parameters, it is hence a local search algorithm with guarantee to converge to a local optimum.
During the local search process, we observe it converges in 3-10 iterations with 1 minute per image on average.
Algorithm \ref{alg:local_search} shows the overall process.
Figure \ref{fig:local_search} illustrates the local search algorithm.

\begin{figure*}[t]
\centering
\includegraphics[width=0.9\textwidth]{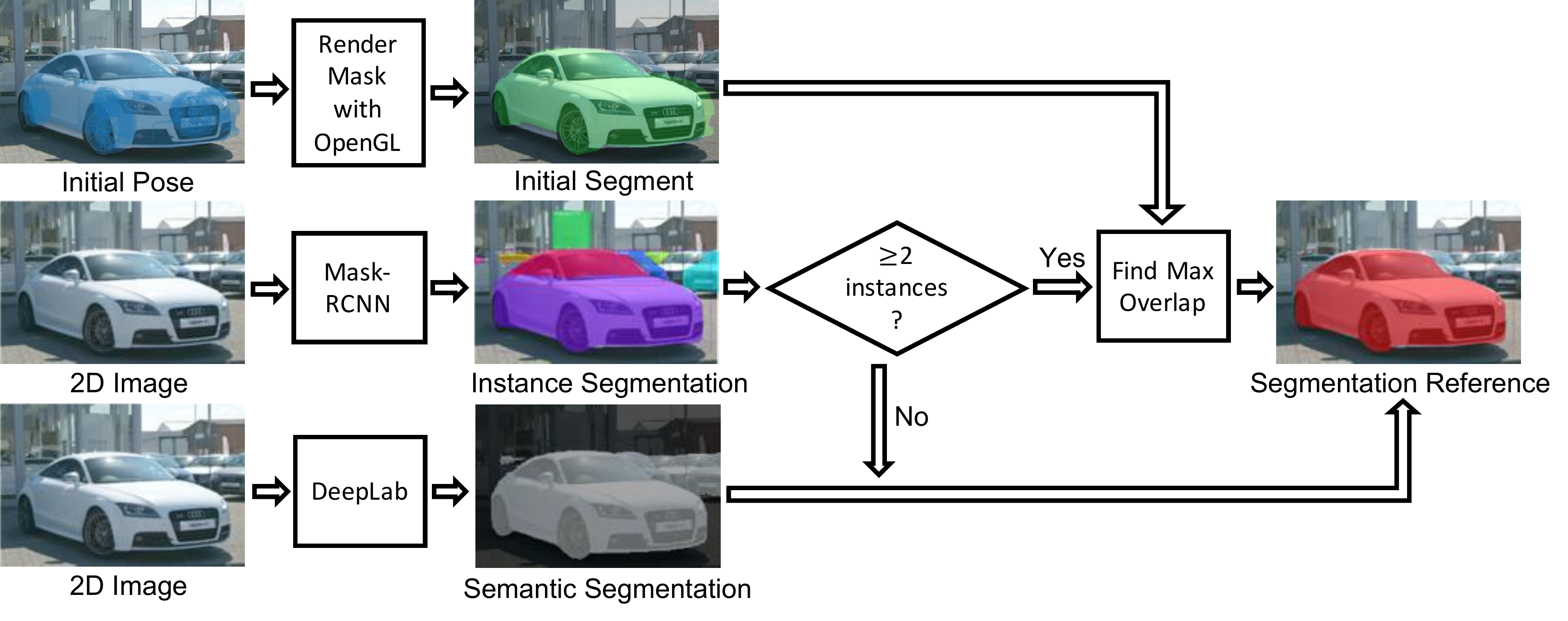}
\caption{An illustration of our reference segmentation extraction process.
Ideally, we can ask human annotators to annotate the ground truth segment for the target object in a 2D image.
However, we find CNNs such as Mask-RCNN and DeepLab can already provide accurate enough segmentation predictions for the pose refinement.
}
\label{fig:extract_segment}
\end{figure*}

\setlength{\tabcolsep}{1pt}
\begin{figure*}[t]
\scriptsize
\centering
\begin{tabular}{c c c c c c c c}
\rotatebox{90}{\hspace{2mm} \scriptsize StanfordCars3D} &
\includegraphics[width=0.12\linewidth]{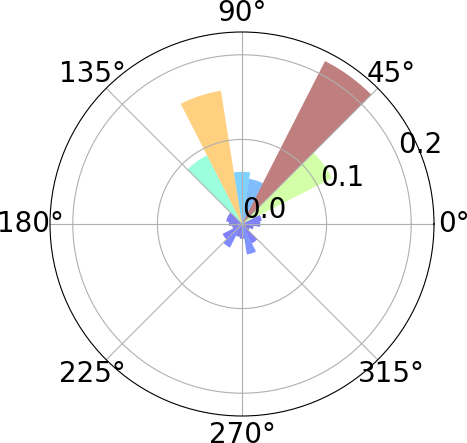} &
\includegraphics[width=0.12\linewidth]{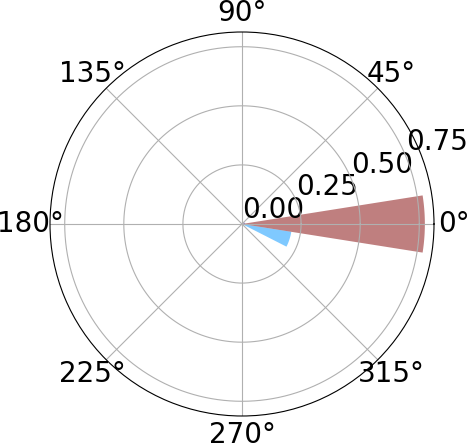} &
\includegraphics[width=0.12\linewidth]{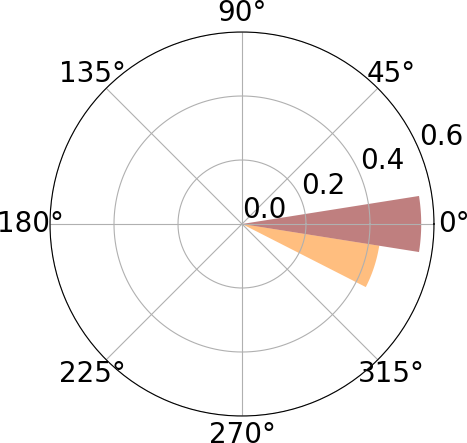} &
\includegraphics[width=0.14\linewidth]{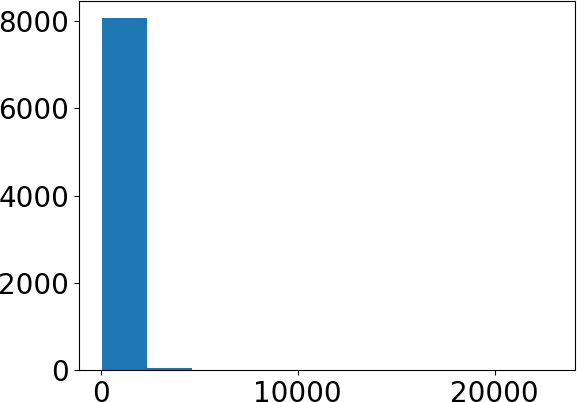} &
\includegraphics[width=0.14\linewidth]{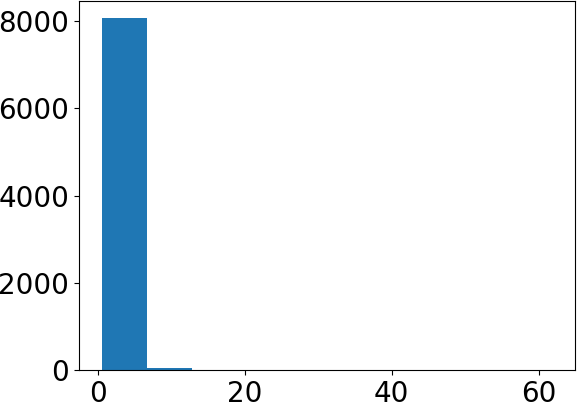} &
\includegraphics[width=0.14\linewidth]{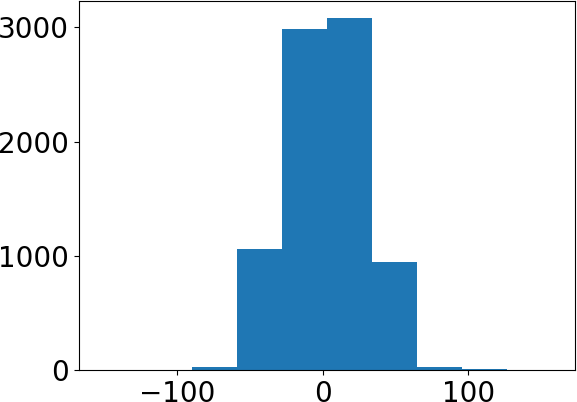} &
\includegraphics[width=0.14\linewidth]{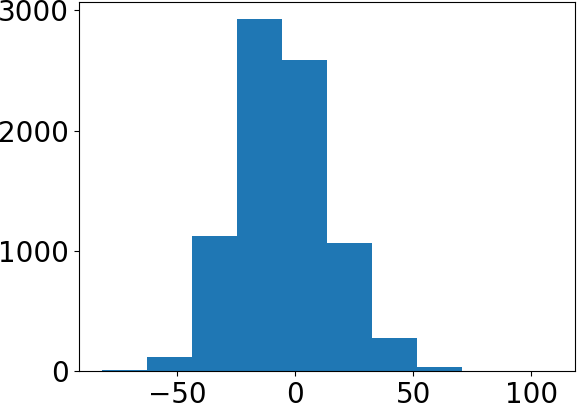} \\
\rotatebox{90}{\hspace{4mm} \scriptsize CompCars3D} &
\includegraphics[width=0.12\linewidth]{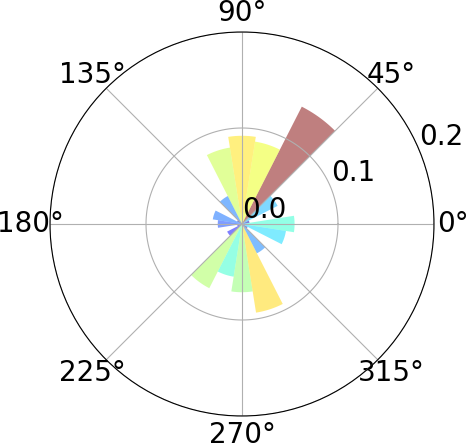} &
\includegraphics[width=0.12\linewidth]{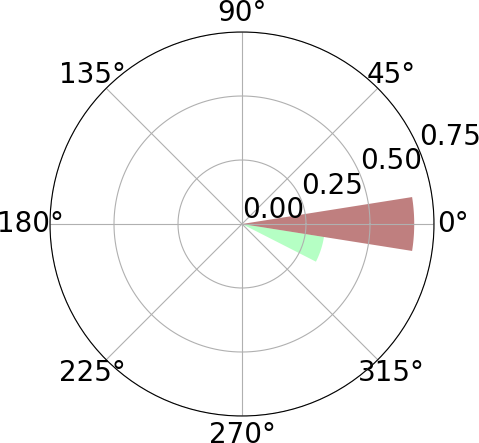} &
\includegraphics[width=0.12\linewidth]{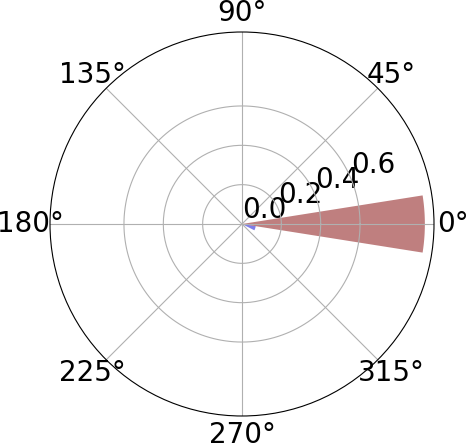} &
\includegraphics[width=0.14\linewidth]{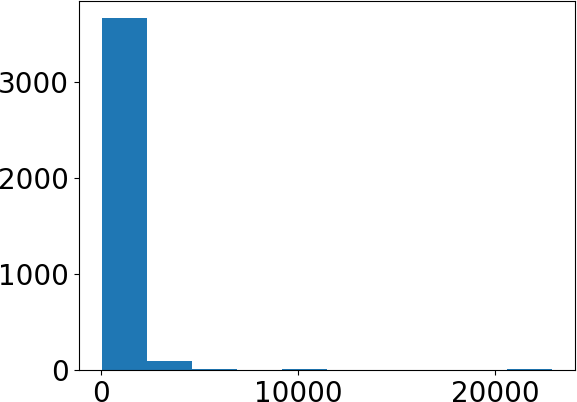} &
\includegraphics[width=0.14\linewidth]{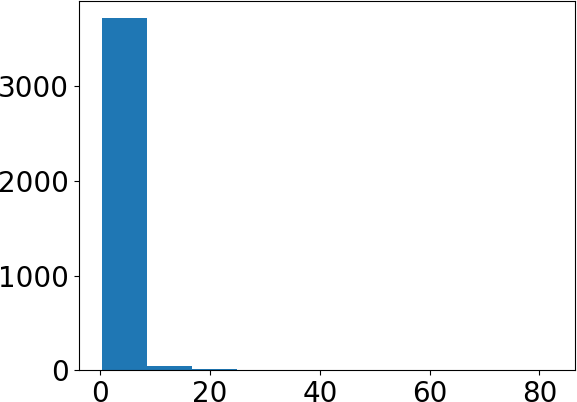} &
\includegraphics[width=0.14\linewidth]{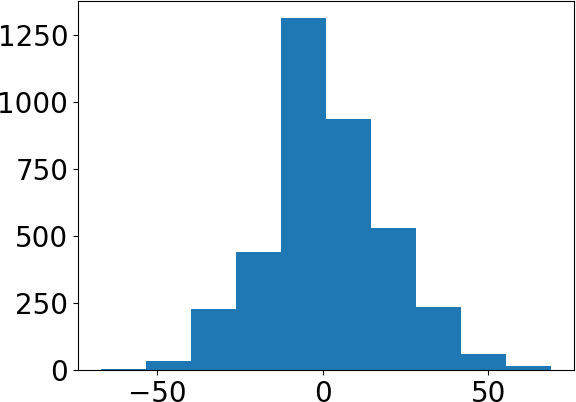} &
\includegraphics[width=0.14\linewidth]{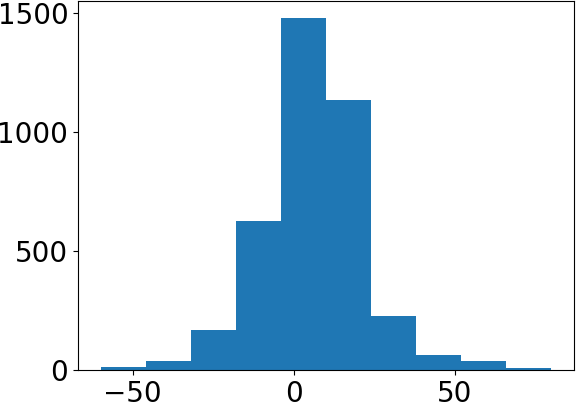} \\
\rotatebox{90}{\hspace{0mm} \scriptsize FGVC-Aircraft3D} &
\includegraphics[width=0.12\linewidth]{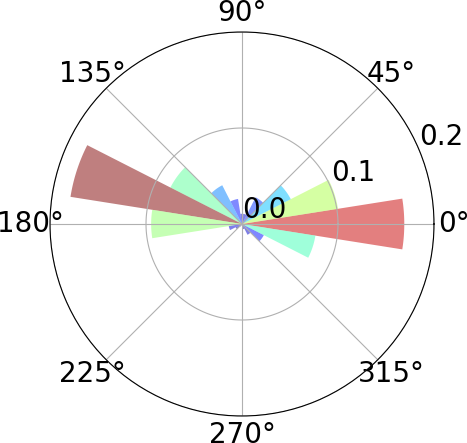} &
\includegraphics[width=0.12\linewidth]{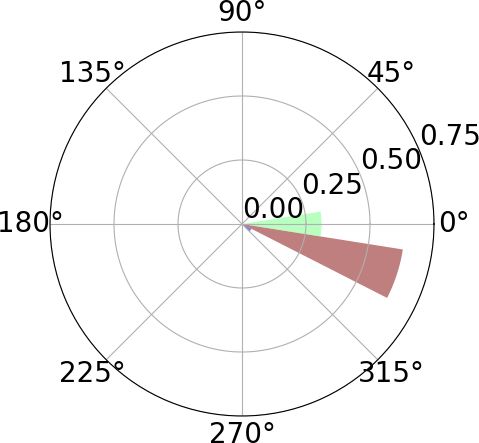} &
\includegraphics[width=0.12\linewidth]{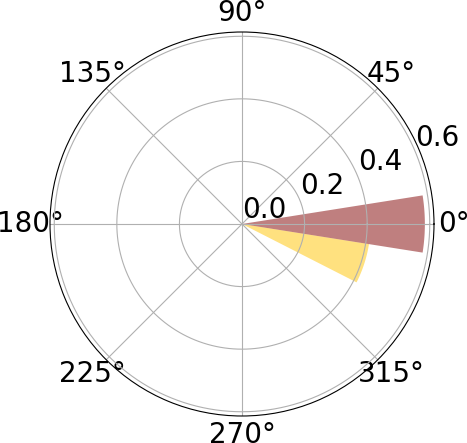} &
\includegraphics[width=0.14\linewidth]{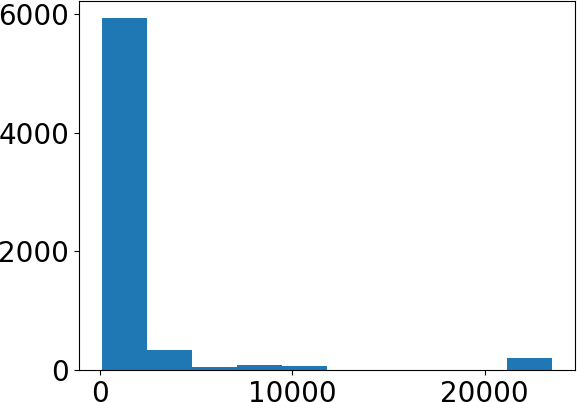} &
\includegraphics[width=0.14\linewidth]{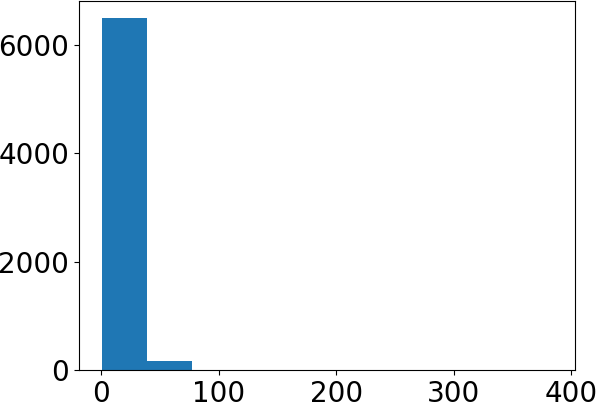} &
\includegraphics[width=0.14\linewidth]{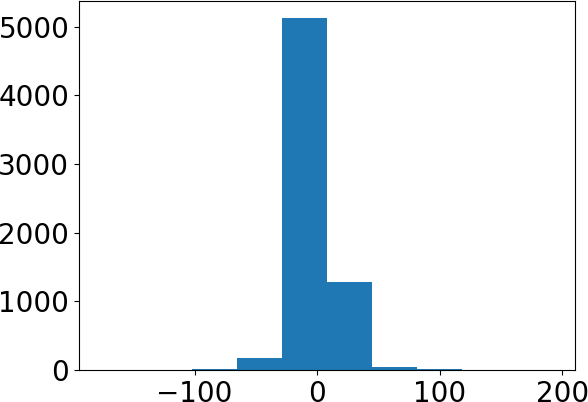} &
\includegraphics[width=0.14\linewidth]{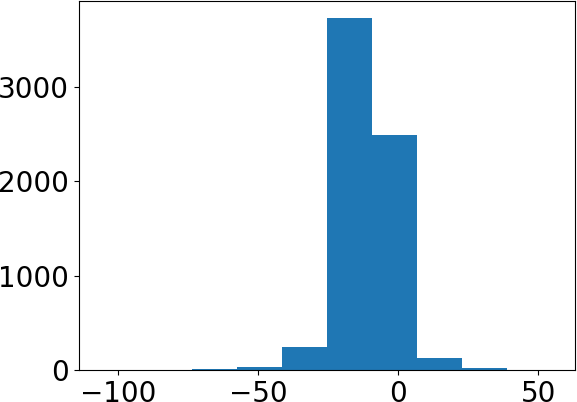} \\
& azimuth $a$ & elevation $e$ & in-plane rotation $\theta$ & focal length $f$ & model depth $d$ & principal $u$ & principal $v$ \\
\end{tabular}
\caption{The polar histograms of the three rotation parameters as well as the histograms of the other four parameters in our annotated dataset.}
\label{fig:statistics}
\end{figure*}

\subsection{Segmentation Reference}

To conduct the local greedy search, ideally we need the ground truth target object segmentation.
Although we may setup another segmentation annotation interface for all 2D images in three datasets through crowd-sourcing, we find using existing state-of-the-art image segmentation models such as Mask R-CNN~\cite{he2017mask} and DeepLab v3+~\cite{deeplabv3plus2018} can already provide us with satisfying segmentation results.
For example, on the Pascal VOC2012 segmentation benchmark, DeepLab v3+ can reach average IoUs of 93.2 on the ``car'' class and 97.0 on the ``aeroplane'' class respectively.
Mask R-CNN, although does not provide as-accurate-as-enough semantic segmentation, is able to obtain instance-level segmentation, which are particularly useful for images with more than 1 instance from the same class.
In the end, we use a combination of both models to find the most appropriate segmentation reference.
Figure \ref{fig:extract_segment} illustrates the process.

\subsection{Dataset Statistics}

We plot the distributions of the 7 parameters in Figure \ref{fig:statistics} for StanfordCars3D, CompCars3D and FGVC-Aircraft3D respectively. 
Due to the nature of the original fine-grained dataset, all the parameters are not uniformly distributed.
Unsurprisingly, the most challenging parameter across the three datasets is azimuth ($a$), which varies across the $360^{\circ}$, while elevation ($e$) and in-plane rotation ($\theta$) are somewhat concentrated in a small range around $0^{\circ}$
since the images of cars and airplanes are often taken from the ground view.
The distribution of focal length ($f$) and model depth ($d$) are also not widespread because objects in these fine-grained images are generally normalized and cropped to a standard size.
Although the parameter distribution issue may raise concerns about learning trivial solutions, we believe that our first attempt still provide reasonable diversity on pose annotation.
For example, the distribution of azimuth ($a$) are quite different across the three datasets and complementary to each other.
This could encourage building a more generalized pose estimation model.

\subsection{Dataset Split}
We split the three datasets in this way.
For StanfordCars3D, since we have annotated all the images, we follow the standard train/test split provided by the original dataset~\cite{krause20133d} with 8144 training examples and 8041 testing examples. For CompCars3D, we randomly sample $2/3$ of our annotated data as training set and the rest $1/3$ as testing set, resulting in 3798 training and 1898 test examples. We provide the train/test split information in the dataset release. For FGVC-Aircraft3D, we follow the standard train/test split provided by the original dataset~\cite{maji2013fine} with 6667 training examples and 3333 testing examples.

\section{Dataset Comparison}

\setlength{\tabcolsep}{1pt}
\begin{figure*}[t]
  \scriptsize
  \centering
  \begin{tabular}{c c c c c c}
    \rotatebox{90}{\hspace{4mm}Pascal3D+} &
    \includegraphics[width=0.16\linewidth,height=0.11\linewidth]{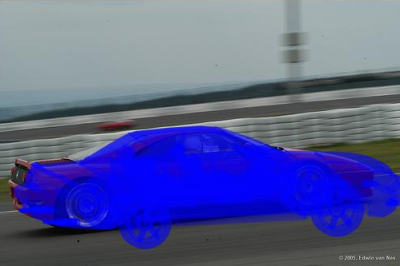} &
    \includegraphics[width=0.16\linewidth,height=0.11\linewidth]{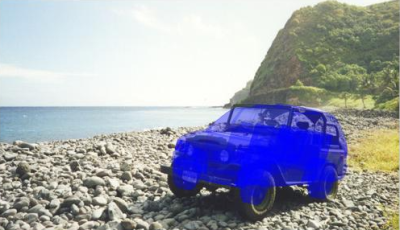} &
    \includegraphics[width=0.16\linewidth,height=0.11\linewidth]{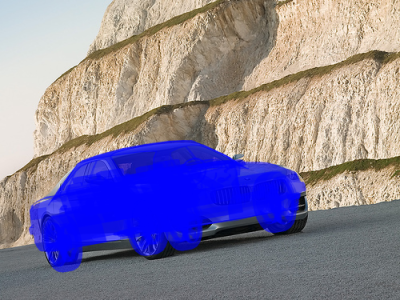} &
    \includegraphics[width=0.16\linewidth,height=0.11\linewidth]{images/car_compare/pascal3d_4.png} &
    \includegraphics[width=0.16\linewidth,height=0.11\linewidth]{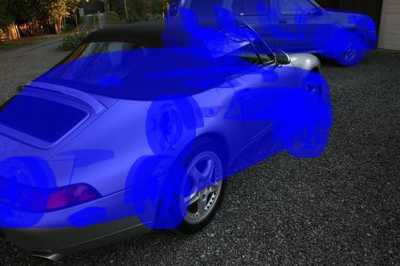} \\
    \rotatebox{90}{\hspace{3mm}ObjectNet3D} & 
    \includegraphics[width=0.16\linewidth,height=0.11\linewidth]{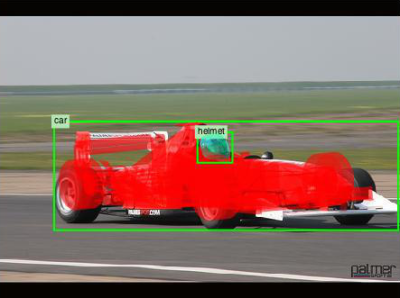} &
    \includegraphics[width=0.16\linewidth,height=0.11\linewidth]{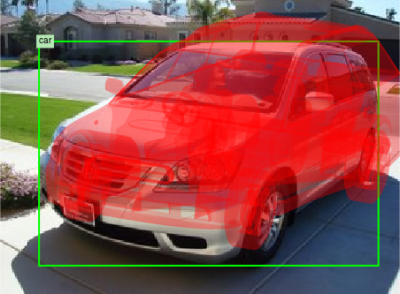} &
    \includegraphics[width=0.16\linewidth,height=0.11\linewidth]{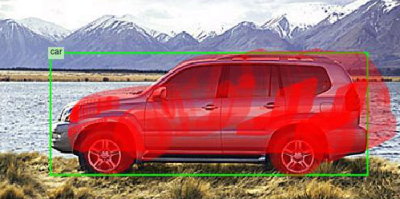} &
    \includegraphics[width=0.16\linewidth,height=0.11\linewidth]{images/car_compare/objnet3d_4.png} & 
    \includegraphics[width=0.16\linewidth,height=0.11\linewidth]{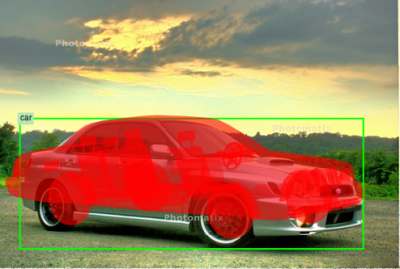} \\
    \rotatebox{90}{\hspace{2mm}StanfordCars3D} &
    \includegraphics[width=0.16\linewidth,height=0.11\linewidth]{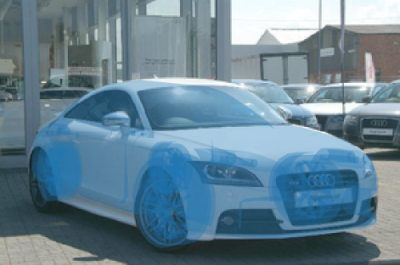} &
    \includegraphics[width=0.16\linewidth,height=0.11\linewidth]{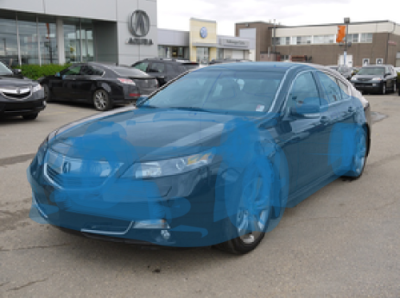} &
    \includegraphics[width=0.16\linewidth,height=0.11\linewidth]{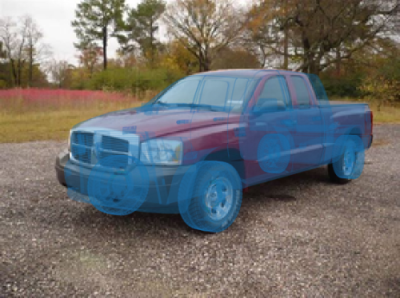} &
    \includegraphics[width=0.16\linewidth,height=0.11\linewidth]{images/car_compare/stanfordcar_4.png} & 
    \includegraphics[width=0.16\linewidth,height=0.11\linewidth]{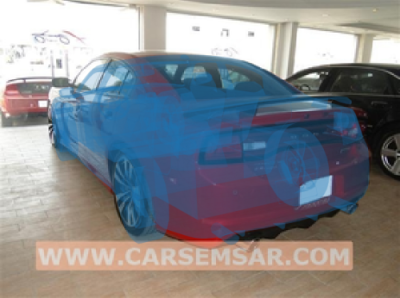} \\
  \end{tabular}
  \caption{Qualitative comparison of ground-truth pose annotation between our StanfordCars3D and two existing large scale 3D pose datatset. We randomly select 5 car images from each dataset. While both Pascal3D+ and ObjectNet3D provide more complicated scenarios with more generic categories for 3D pose estimation, our pose annotations look more accurate thanks to the fine-grained shape matching.}
  \label{fig:car_compare}
\end{figure*}

\setlength{\tabcolsep}{1pt}
\begin{figure*}[t]
  \scriptsize
  \centering
  \begin{tabular}{c c c c c c}
    \rotatebox{90}{\hspace{4mm}Pascal3D+} &
    \includegraphics[width=0.16\linewidth,height=0.11\linewidth]{images/aircraft_compare/pascal3d_1.png} &
    \includegraphics[width=0.16\linewidth,height=0.11\linewidth]{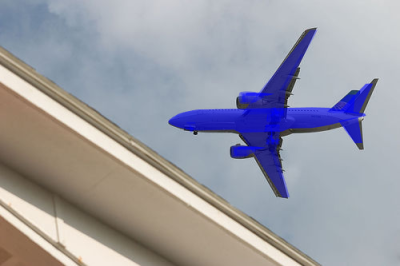} &
    \includegraphics[width=0.16\linewidth,height=0.11\linewidth]{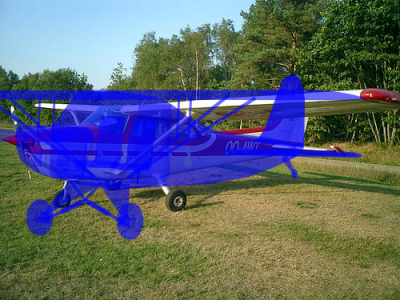} &
    \includegraphics[width=0.16\linewidth,height=0.11\linewidth]{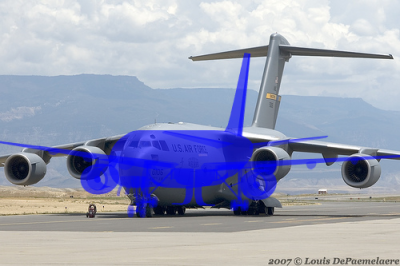} &
    \includegraphics[width=0.16\linewidth,height=0.11\linewidth]{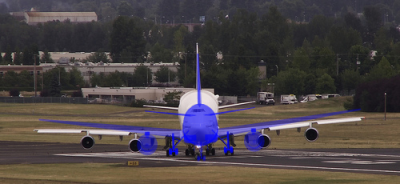} \\
    \rotatebox{90}{\hspace{3mm}ObjectNet3D} & 
    \includegraphics[width=0.16\linewidth,height=0.11\linewidth]{images/aircraft_compare/objnet3d_1.png} &
    \includegraphics[width=0.16\linewidth,height=0.11\linewidth]{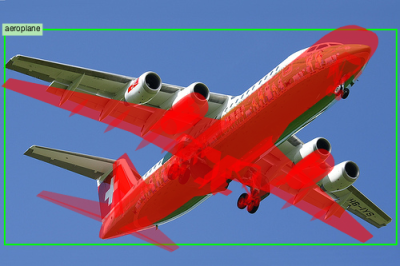} &
    \includegraphics[width=0.16\linewidth,height=0.11\linewidth]{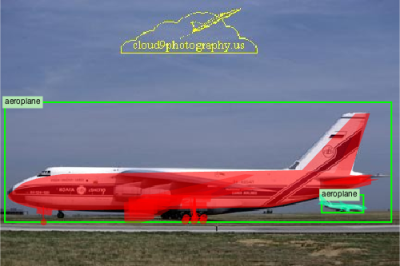} &
    \includegraphics[width=0.16\linewidth,height=0.11\linewidth]{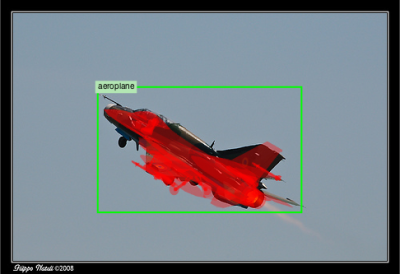} & 
    \includegraphics[width=0.16\linewidth,height=0.11\linewidth]{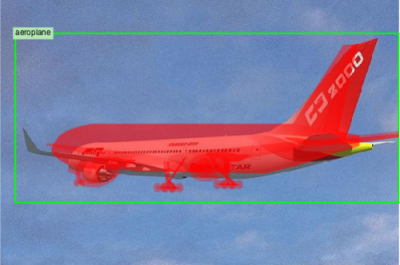} \\
    \rotatebox{90}{\hspace{0.5mm}FGVC-Aircraft3D} &
    \includegraphics[width=0.16\linewidth,height=0.11\linewidth]{images/aircraft_compare/fgvc_1.png} &
    \includegraphics[width=0.16\linewidth,height=0.11\linewidth]{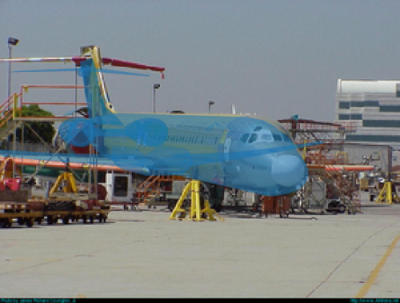} &
    \includegraphics[width=0.16\linewidth,height=0.11\linewidth]{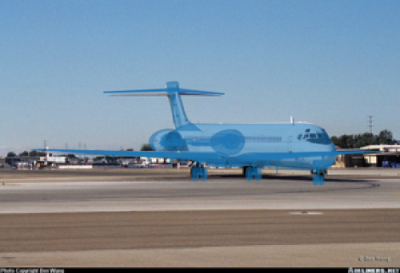} &
    \includegraphics[width=0.16\linewidth,height=0.11\linewidth]{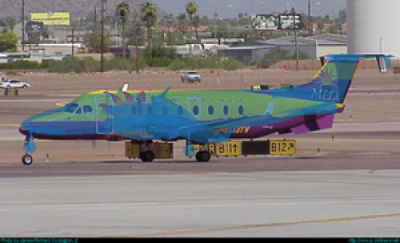} & 
    \includegraphics[width=0.16\linewidth,height=0.11\linewidth]{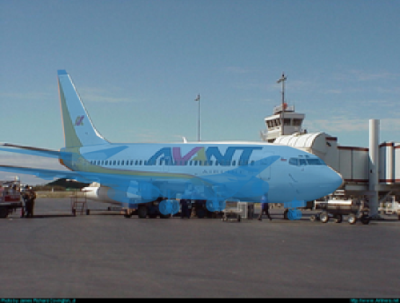} \\
  \end{tabular}
  \caption{Qualitative comparison of ground-truth pose annotation between our FGVC-Aircraft3D and two existing large scale 3D pose datatset. We randomly select 5 aircraft images from each dataset.}
  \label{fig:aircraft_compare}
\end{figure*}

\setlength{\tabcolsep}{1pt}
\begin{figure*}[t]
  \scriptsize
  \centering
  \begin{tabular}{c c c c c c c}
    \rotatebox{90}{\hspace{4mm}Initial Pose} &
    \includegraphics[width=0.16\linewidth,height=0.11\linewidth]{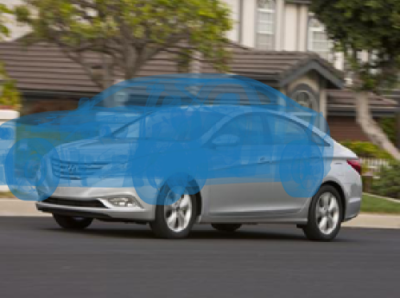} &
    \includegraphics[width=0.16\linewidth,height=0.11\linewidth]{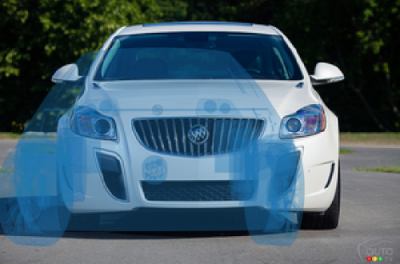} &
    \includegraphics[width=0.16\linewidth,height=0.11\linewidth]{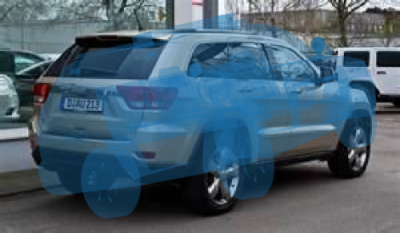} &
    \includegraphics[width=0.16\linewidth,height=0.11\linewidth]{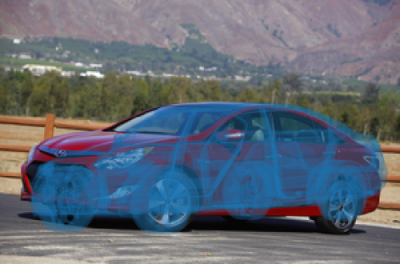} &
    \includegraphics[width=0.16\linewidth,height=0.11\linewidth]{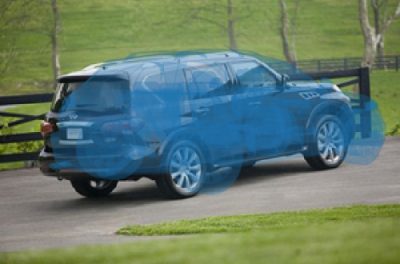} \\
    \rotatebox{90}{\hspace{4mm}Final Pose} & 
    \includegraphics[width=0.16\linewidth,height=0.11\linewidth]{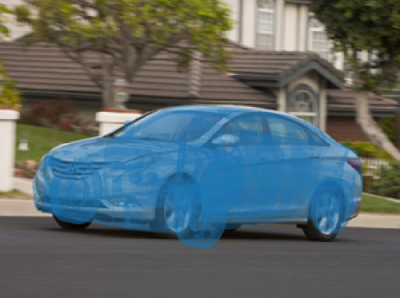} &
    \includegraphics[width=0.16\linewidth,height=0.11\linewidth]{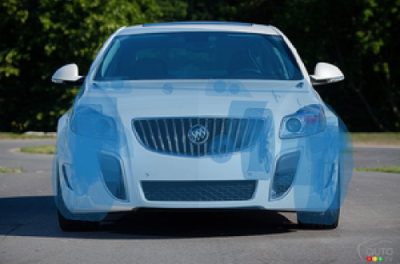} &
    \includegraphics[width=0.16\linewidth,height=0.11\linewidth]{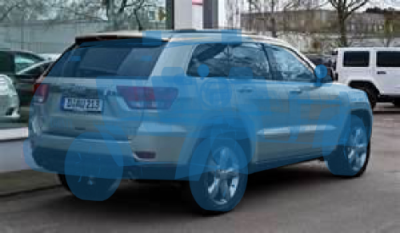} &
    \includegraphics[width=0.16\linewidth,height=0.11\linewidth]{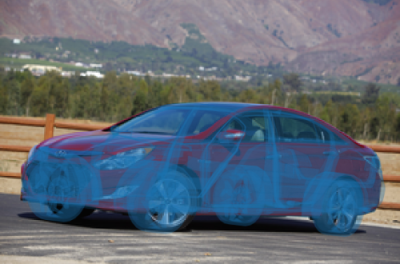} & 
    \includegraphics[width=0.16\linewidth,height=0.11\linewidth]{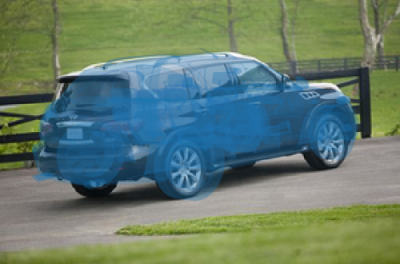} \\
  \end{tabular}
  \caption{Selected examples illustrating the second-stage automatic pose refinement improving the initial human pose annotation on StanfordCars3D dataset.}
  \label{fig:car_version_compare}
\end{figure*}

\setlength{\tabcolsep}{1pt}
\begin{figure*}[t]
  \scriptsize
  \centering
  \begin{tabular}{c c c c c c c}
    \rotatebox{90}{\hspace{4mm}Initial Pose} &
    \includegraphics[width=0.16\linewidth,height=0.11\linewidth]{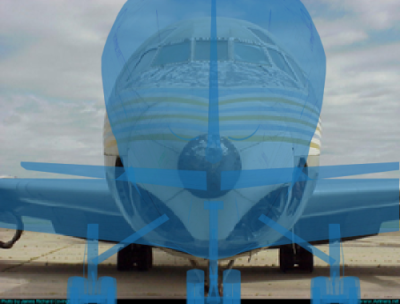} &
    \includegraphics[width=0.16\linewidth,height=0.11\linewidth]{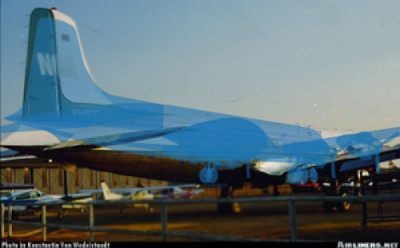} &
    \includegraphics[width=0.16\linewidth,height=0.11\linewidth]{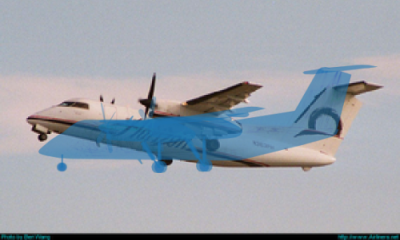} &
    \includegraphics[width=0.16\linewidth,height=0.11\linewidth]{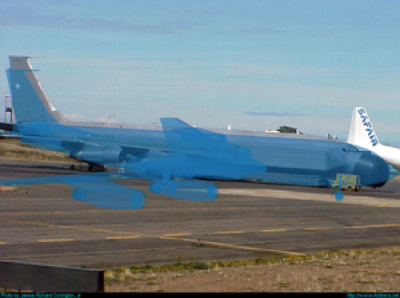} &
    \includegraphics[width=0.16\linewidth,height=0.11\linewidth]{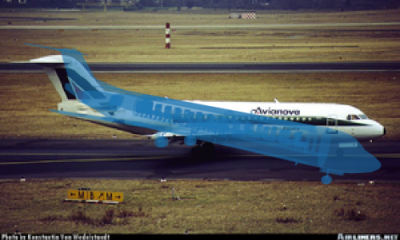} \\
    \rotatebox{90}{\hspace{4mm}Final Pose} & 
    \includegraphics[width=0.16\linewidth,height=0.11\linewidth]{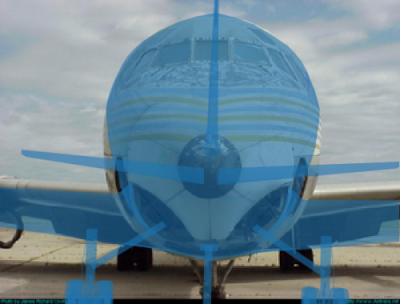} &
    \includegraphics[width=0.16\linewidth,height=0.11\linewidth]{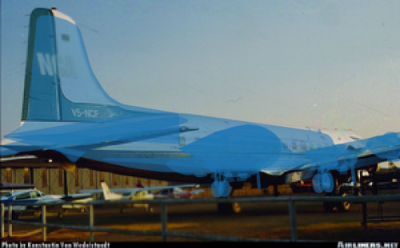} &
    \includegraphics[width=0.16\linewidth,height=0.11\linewidth]{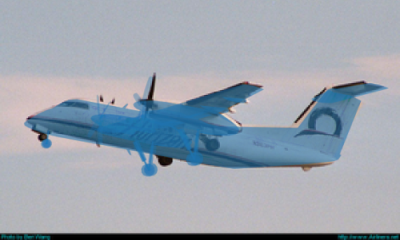} &
    \includegraphics[width=0.16\linewidth,height=0.11\linewidth]{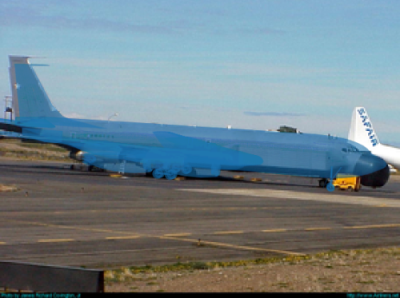} & 
    \includegraphics[width=0.16\linewidth,height=0.11\linewidth]{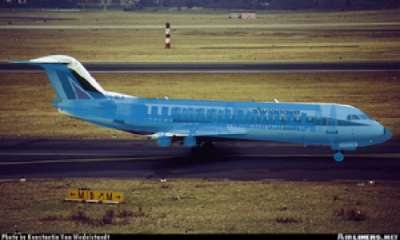} \\
  \end{tabular}
  \caption{Selected examples illustrating the second-stage automatic pose refinement improving the initial human pose annotation on FGVC-Aircraft3D dataset.}
  \label{fig:aircraft_version_compare}
\end{figure*}

\subsection{Compare to Existing Dataset}
We compare our annotation quality with two existing large-scale 3D pose dataset, PASCAL3D+~\cite{xiang2014beyond} and ObjectNet3D~\cite{xiang2016objectnet3d}.
It is worth to note that we are not aiming to show the superiority of our dataset, since both previous datasets consider more general scenarios with multiple objects and challenging occlusion in an image.
However, we hope that by comparing to them, we demonstrate our fine-grained pose dataset can become a complementary resource for studying 3D pose estimation in monocular images.

Figure~\ref{fig:car_compare} and Figure~\ref{fig:aircraft_compare} show the qualitative comparison on the ``car'' class and the ``aeroplane'' class respectively.
Overall, we find our annotation more satisfying by visually comparing the overlay images which maps the 3D model on the 2D image.
To further conduct quantitative comparison, we use segmentation overlap between the projected 2D mask and the ground truth object mask as the evaluation measure.
We randomly select 50 ``car'' images and 50 ``aeroplane'' images from PASCAL3D+ and ObjectNet3D respectively.
We then randomly pick 50 images from StanfordCars3D and FGVC-Aircraft3D.
In total, we randomly select 300 images and annotate them with ground truth segmentation.
Since both PASCAL3D+ and StanfordCars3D consider more complicated scenarios such as multiple objects with cluttered background, we filter out those images containing more than one object with reasonably large size for a fair comparison.
Hence the average IoUs can be an optimistic estimate for both baseline datasets.
Even with that, our annotation shows a clear segmentation improvement on average IoUs on both ``car'' and ``aeroplane'', as demonstrated in Table~\ref{table:iou_compare}.
Particularly, both the mean and the standard deviation of the segmentation IoUs get significantly improved, indictating that our annotations are not only more accurate but more stable as well.

\begin{table}
\small
\centering
\begin{tabular}{|c|c|c|c|}
\hline
car & PASCAL3D+~\cite{xiang2014beyond} & ObjectNet3D~\cite{xiang2016objectnet3d} & StanfordCars3D \\
\hline
& 78.5\% $\pm$ 8.6\% & 84.1\% $\pm$ 6.0\% & {\bf 90.4\% $\pm$ 3.3\%} \\
\hline
airplane & PASCAL3D+~\cite{xiang2014beyond} & ObjectNet3D~\cite{xiang2016objectnet3d} & FGVC-Aircraft3D \\
\hline
& 62.7\% $\pm$ 13.1\% & 65.1\% $\pm$ 11.0\% & {\bf 78.9\% $\pm$ 9.4\%} \\
\hline
\end{tabular}
\caption{Comparison on the average IoUs with the standard deviation on the ``car'' category and ``aeroplane'' category. Note that in this evaluation, we manually annotate around 50 ground truth segmentation masks for each dataset.}
\label{table:iou_compare}
\end{table}


\subsection{Compare to Human Annotation}
We also analyze how much gain we get by conducting segmentation based pose refinement.
To understand this, we utilize the manually annotated ground truth 2D segmentation on the randomly select 100 images from the StanfordCars and FGVC-Aircraft.
We then compare the average IoUs between human annotated pose and the refined pose.
Table~\ref{table:version_iou_compare} shows the improvement of segmentation overlap on the three datasets.
On StanfordCars3D, for example, our second-stage refinement improves average IoUs from 84.1\% to 90.4\%, which is significant.
On FGVC-Aircarft3D, the improvement is even more, from 65.3\% to 78.9\%.
Figure~\ref{fig:car_version_compare} and Figure~\ref{fig:aircraft_version_compare} illustrate the pose improvement qualitatively.

Considering segmentation overlap may not be the only appropriate quantitative measure, we further conduct a human study to compare the pose annotation quality.
To do this, we hire 5 professional annotators, show them the 2D-3D alignment of the same image with annotation in the two stages simultaneously and let them rate the relative quality for the 50 selected images in each dataset.
The relative comparison consists of ``Worse'', ``Equal'' and ``Better'', indicating the second-stage pose is either significantly worse, roughly equal or significantly better than the first-stage human annotation from the subjective point of view.
Table~\ref{table:version_human_study} shows the human study result.
Most of the time, the second-stage refined pose is either roughly equal or significantly better than the initial human annotation, suggesting the benefit of utilizing segmentation cues to facilitate the pose search.

\begin{table}
\small
\centering
\begin{tabular}{|c|c|c|}
\hline
Average IoUs & Human Annotation & Refined Annotation \\
\hline
StanfordCars3D & 84.1\% $\pm$ 6.2\% & {\bf 90.4\% $\pm$ 3.3\%} \\
\hline
FGVC-Aircraft3D & 65.3\% $\pm$ 19.9\% & {\bf 78.9\% $\pm$ 9.4\%} \\
\hline
\end{tabular}
\caption{Segmentation evaluation of initial human annotation and after iterative pose refinement on the two datasets. Note that in this evaluation, we manually annotate around 50 ground truth segmentation masks for each dataset.}
\label{table:version_iou_compare}
\end{table}

\begin{table}
\small
\centering
\begin{tabular}{|c|c|c|c|}
\hline
 & Worse & Equal & Better \\
\hline
StanfordCars3D & 13.0\% & 28.3\% & 58.7\% \\
\hline
FGVC-Aircraft3D & 12.8\% & 40.4\% & 46.8\% \\
\hline
\end{tabular}
\caption{Human satisfaction rate by comparing original human annotation with refined pose. ``Worse'' means refined pose is worse than initial pose. ``Better'' means refined is better. ``Equal'' meaning the annotation are roughly the same. From the table, we can see humans are much more satisfied with the refined pose annotation.}
\label{table:version_human_study}
\end{table}

\section{Discussions}
In summary, we introduce the new problem of 3D pose estimation for fine-grained object categories from a monocular image 
We annotate three popular fine-grained recognition datasets with 3D shapes and poses, ending in total 31,881 images with 409 classes.
By utilizing image segmentation as an intermediate cue, we further improve the pose annotation quality.
It is worth to note that human may ultimately produce better annotation given unlimited time, but the segmentation based pose refinement provides a facilitation with a better trade-off between cost and accuracy.

There are still a need of future works to continue the improvement.
First, the super-categories shall be continued to enlarge with more fine-grained datasets.
Second, the current fine-grained datasets are less challenging in terms of background clutter and object size.
Third, while all existing large-scale pose datasets limit to rigid objects, it is still necessary to develop methods for non-rigid objects.
Finally, it is also possible to develop a neural network architecture to replace the segmentation based pose refinement and combine it with the human annotation interface.
We leave these as future work.

{\small
\bibliographystyle{ieee}
\bibliography{egbib}

\begin{thebibliography}{10}\itemsep=-1pt

\bibitem{chang2015shapenet}
A.~X. Chang, T.~Funkhouser, L.~Guibas, P.~Hanrahan, Q.~Huang, Z.~Li,
  S.~Savarese, M.~Savva, S.~Song, H.~Su, et~al.
\newblock {ShapeNet}: An information-rich {3D} model repository.
\newblock {\em arXiv preprint arXiv:1512.03012}, 2015.

\bibitem{deeplabv3plus2018}
L.-C. Chen, Y.~Zhu, G.~Papandreou, F.~Schroff, and H.~Adam.
\newblock Encoder-decoder with atrous separable convolution for semantic image
  segmentation.
\newblock In {\em ECCV}, 2018.

\bibitem{chen2009benchmark}
X.~Chen, A.~Golovinskiy, and T.~Funkhouser.
\newblock A benchmark for 3d mesh segmentation.
\newblock In {\em Acm transactions on graphics (tog)}, volume~28, page~73. ACM,
  2009.

\bibitem{chen2016monocular}
X.~Chen, K.~Kundu, Z.~Zhang, H.~Ma, S.~Fidler, and R.~Urtasun.
\newblock Monocular {3D} object detection for autonomous driving.
\newblock In {\em CVPR}, 2016.

\bibitem{chen2012schelling}
X.~Chen, A.~Saparov, B.~Pang, and T.~Funkhouser.
\newblock Schelling points on 3d surface meshes.
\newblock {\em ACM Transactions on Graphics (TOG)}, 31(4):29, 2012.

\bibitem{everingham2010pascal}
M.~Everingham, L.~Van~Gool, C.~K. Williams, J.~Winn, and A.~Zisserman.
\newblock The pascal visual object classes (voc) challenge.
\newblock {\em International journal of computer vision}, 88(2):303--338, 2010.

\bibitem{ghodrati20142d}
A.~Ghodrati, M.~Pedersoli, and T.~Tuytelaars.
\newblock Is {2D} information enough for viewpoint estimation?
\newblock In {\em BMVC}, 2014.

\bibitem{he2017mask}
K.~He, G.~Gkioxari, P.~Doll{\'a}r, and R.~Girshick.
\newblock Mask {R-CNN}.
\newblock In {\em ICCV}, 2017.

\bibitem{jayawardena2013image}
S.~Jayawardena et~al.
\newblock {\em Image based automatic vehicle damage detection}.
\newblock PhD thesis, The Australian National University, 2013.

\bibitem{kim2013learning}
V.~G. Kim, W.~Li, N.~J. Mitra, S.~Chaudhuri, S.~DiVerdi, and T.~Funkhouser.
\newblock Learning part-based templates from large collections of 3d shapes.
\newblock {\em ACM Transactions on Graphics (TOG)}, 32(4):70, 2013.

\bibitem{krause2016unreasonable}
J.~Krause, B.~Sapp, A.~Howard, H.~Zhou, A.~Toshev, T.~Duerig, J.~Philbin, and
  L.~Fei-Fei.
\newblock The unreasonable effectiveness of noisy data for fine-grained
  recognition.
\newblock In {\em European Conference on Computer Vision}, pages 301--320.
  Springer, 2016.

\bibitem{krause20133d}
J.~Krause, M.~Stark, J.~Deng, and L.~Fei-Fei.
\newblock {3D} object representations for fine-grained categorization.
\newblock In {\em ICCV Workshops on 3D Representation and Recognition}, 2013.

\bibitem{krizhevsky2012imagenet}
A.~Krizhevsky, I.~Sutskever, and G.~E. Hinton.
\newblock {ImageNet} classification with deep convolutional neural networks.
\newblock In {\em NIPS}, 2012.

\bibitem{li2014shrec}
B.~Li, Y.~Lu, C.~Li, A.~Godil, T.~Schreck, M.~Aono, M.~Burtscher, H.~Fu,
  T.~Furuya, H.~Johan, et~al.
\newblock Shrec’14 track: Extended large scale sketch-based 3d shape
  retrieval.
\newblock In {\em Eurographics workshop on 3D object retrieval}, volume 2014.
  ., 2014.

\bibitem{lim2013parsing}
J.~J. Lim, H.~Pirsiavash, and A.~Torralba.
\newblock Parsing {IKEA} objects: Fine pose estimation.
\newblock In {\em ICCV}, 2013.

\bibitem{lin2015bilinear}
T.-Y. Lin, A.~RoyChowdhury, and S.~Maji.
\newblock Bilinear {CNN} models for fine-grained visual recognition.
\newblock In {\em ICCV}, 2015.

\bibitem{mahendran20173d}
S.~Mahendran, H.~Ali, and R.~Vidal.
\newblock 3d pose regression using convolutional neural networks.
\newblock In {\em IEEE International Conference on Computer Vision}, volume~1,
  page~4, 2017.

\bibitem{maji2013fine}
S.~Maji, E.~Rahtu, J.~Kannala, M.~Blaschko, and A.~Vedaldi.
\newblock Fine-grained visual classification of aircraft.
\newblock {\em arXiv preprint arXiv:1306.5151}, 2013.

\bibitem{mottaghi2015coarse}
R.~Mottaghi, Y.~Xiang, and S.~Savarese.
\newblock A coarse-to-fine model for {3D} pose estimation and sub-category
  recognition.
\newblock In {\em CVPR}, 2015.

\bibitem{ozuysal2009pose}
M.~Ozuysal, V.~Lepetit, and P.~Fua.
\newblock Pose estimation for category specific multiview object localization.
\newblock In {\em CVPR}, 2009.

\bibitem{park2017transformation}
E.~Park, J.~Yang, E.~Yumer, D.~Ceylan, and A.~C. Berg.
\newblock Transformation-grounded image generation network for novel {3D} view
  synthesis.
\newblock In {\em CVPR}, 2017.

\bibitem{savarese20073d}
S.~Savarese and L.~Fei-Fei.
\newblock {3D} generic object categorization, localization and pose estimation.
\newblock In {\em ICCV}, 2007.

\bibitem{shilane2004princeton}
P.~Shilane, P.~Min, M.~Kazhdan, and T.~Funkhouser.
\newblock The princeton shape benchmark.
\newblock In {\em Shape modeling applications, 2004. Proceedings}, pages
  167--178. IEEE, 2004.

\bibitem{sochor2016boxcars}
J.~Sochor, A.~Herout, and J.~Havel.
\newblock {BoxCars}: {3D} boxes as cnn input for improved fine-grained vehicle
  recognition.
\newblock In {\em CVPR}, 2016.

\bibitem{van2017inaturalist}
G.~Van~Horn, O.~Mac~Aodha, Y.~Song, A.~Shepard, H.~Adam, P.~Perona, and
  S.~Belongie.
\newblock The inaturalist challenge 2017 dataset.
\newblock {\em arXiv preprint arXiv:1707.06642}, 2017.

\bibitem{varley2017shape}
J.~Varley, C.~DeChant, A.~Richardson, J.~Ruales, and P.~Allen.
\newblock Shape completion enabled robotic grasping.
\newblock In {\em IROS}, 2017.

\bibitem{wah2011cub}
C.~Wah, S.~Branson, P.~Welinder, P.~Perona, and S.~Belongie.
\newblock {The Caltech-UCSD Birds-200-2011 Dataset}.
\newblock Technical Report CNS-TR-2011-001, California Institute of Technology,
  2011.

\bibitem{xiang2015data}
Y.~Xiang, W.~Choi, Y.~Lin, and S.~Savarese.
\newblock Data-driven 3d voxel patterns for object category recognition.
\newblock In {\em Proceedings of the IEEE Conference on Computer Vision and
  Pattern Recognition}, pages 1903--1911, 2015.

\bibitem{xiang2016objectnet3d}
Y.~Xiang, W.~Kim, W.~Chen, J.~Ji, C.~Choy, H.~Su, R.~Mottaghi, L.~Guibas, and
  S.~Savarese.
\newblock {ObjectNet3D}: A large scale database for {3D} object recognition.
\newblock In {\em ECCV}, 2016.

\bibitem{xiang2014beyond}
Y.~Xiang, R.~Mottaghi, and S.~Savarese.
\newblock Beyond {PASCAL}: A benchmark for {3D} object detection in the wild.
\newblock In {\em WACV}, 2014.

\bibitem{yang2015large}
L.~Yang, P.~Luo, C.~Change~Loy, and X.~Tang.
\newblock A large-scale car dataset for fine-grained categorization and
  verification.
\newblock In {\em CVPR}, 2015.

\bibitem{zhou2016view}
T.~Zhou, S.~Tulsiani, W.~Sun, J.~Malik, and A.~A. Efros.
\newblock View synthesis by appearance flow.
\newblock In {\em ECCV}, 2016.

\bibitem{zia2013detailed}
M.~Z. Zia, M.~Stark, B.~Schiele, and K.~Schindler.
\newblock Detailed {3D} representations for object recognition and modeling.
\newblock {\em PAMI}, 35(11):2608--2623, 2013.

\end{thebibliography}
}

\end{document}